\pdfoutput=1
\PassOptionsToPackage{dvipsnames,svgnames,table}{xcolor}

\documentclass[11pt]{article}

\usepackage[]{ACL2023}

\usepackage{times}
\usepackage{latexsym}

\usepackage[T1]{fontenc}

\usepackage[utf8]{inputenc}

\usepackage{microtype}

\usepackage{inconsolata}

\usepackage{graphicx} 
\usepackage[table, dvipsnames]{xcolor}
\usepackage{tcolorbox}
\usepackage{amsmath, amssymb}
\usepackage{mathtools}
\usepackage{longtable}
\usepackage{float}
\usepackage{booktabs}    
\usepackage{multirow}

\usepackage{algorithm}      
\usepackage{algpseudocode}
\usepackage{amsfonts}
\usepackage{bm}

\title{SPEAR: Distilling Domain-Adaptive Reasoning Skeletons via Sequential Symbolic Alignment in Reinforcement Learning}

\author{
 \textbf{Zhuochun Li\textsuperscript{1}},
 \textbf{Yuelyu Ji\textsuperscript{1}},
  \textbf{Yiming Zeng\textsuperscript{2}},
 \textbf{Daqing He\textsuperscript{1}},
\\
 \textsuperscript{1}University of Pittsburgh, Pittsburgh, USA \\
 \textsuperscript{2}University of Connecticut, Storrs, USA
\\
   \{zhl163, yuj49, dah44\}@pitt.edu
}

\begin{document}
\maketitle
\begin{abstract}
Reinforcement learning-based knowledge distillation has the potential to transfer complex reasoning from teacher to student models, yet it currently faces a critical dilemma: researchers must choose between sparse outcome-based rewards, which provide insufficient logical guidance, or expensive neural Process Reward Models (PRMs) for dense signals. 
We resolve this by introducing \textbf{SPEAR} (\textbf{S}ymbolic \textbf{P}rocess \textbf{E}valuation and \textbf{A}lignment \textbf{R}eward), a training-free and plug-and-play process reward method for sequence-level on-policy distillation. SPEAR projects natural-language reasoning traces into domain-adaptive symbolic milestones, providing an efficient proxy for process-level reasoning alignment. By utilizing the longest common subsequence (LCS) to align student explorations with teacher milestones, SPEAR provides a dense, order-aware reward signal that aligns logical consistency without the need for an external neural verifier. Our experiments across math, science, and commonsense reasoning tasks demonstrate that SPEAR effectively bridges the reasoning gap between student and teacher models via sequence-level distillation with efficient dense process rewards. 
Our code and data are available at \url{https://github.com/zhuochunli/SPEAR}.

\end{abstract}

\section{Introduction}
Distilling complex reasoning capabilities from proprietary Large Language Models (LLMs) to resource-efficient Small Language Models (SLMs) is essential for practical deployment. Supervised fine-tuning (SFT) on teacher-generated rationales is a common form of off-policy distillation~\cite{magister2023teaching, xu2024survey}, but can create a mismatch between training responses and the student's own generations. On-policy distillation instead applies teacher guidance to student-generated trajectories~\cite{xu2025beyond, li2026rethinking}. This distinction concerns the source of trajectories: distillation can use distribution matching, reinforcement learning (RL), or both. SPEAR studies teacher-guided RL, where symbolic milestones extracted from teacher rationales supervise student rollouts through a sequence-level reward. Within RL, outcome rewards provide limited guidance about intermediate reasoning, while learned Process Reward Models (PRMs) require additional model inference~\cite{lightman2024let}. This motivates an inexpensive reward that incorporates teacher reasoning without requiring a neural reward model or teacher token probabilities.

\begin{figure}[tb]
    \centering
     \includegraphics[width=0.48\textwidth,height=0.28\textwidth]{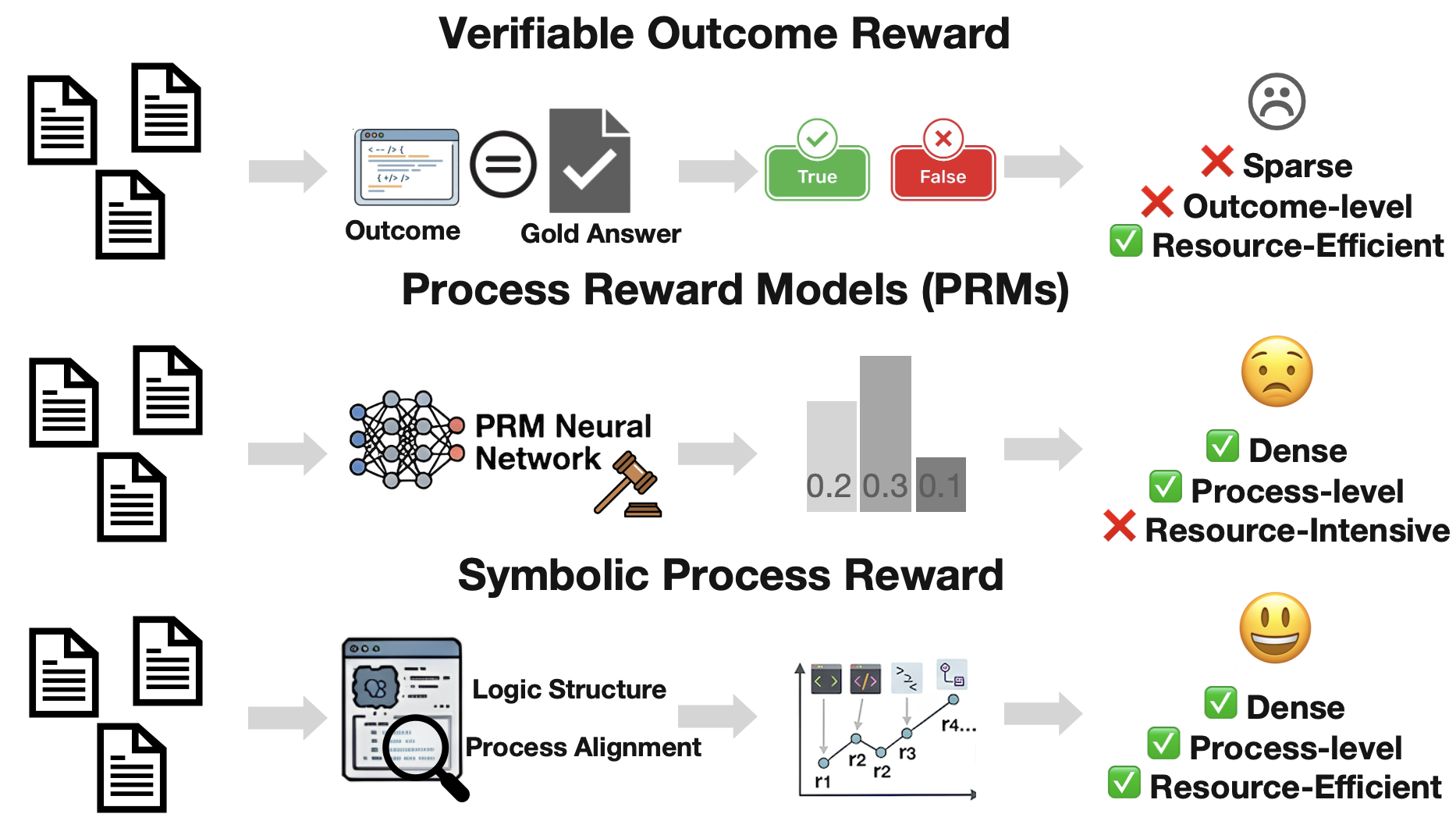}
     \caption{Comparison between SPEAR and other reward methods.}
    \label{fig_pipeline}
    \vspace{-0.45cm}   
\end{figure}

Existing process-supervision methods are largely limited to formal domains such as mathematics and programming, where intermediate steps and final answers are deterministically verifiable~\cite{shao2024deepseekmath, song2025prmbench, yu2024reasoning}. Extending process rewards to broader reasoning tasks (e.g., scientific and commonsense reasoning) remains challenging~\cite{yu2023thought, weng2024leveraging}, as natural language reasoning lacks clear structural boundaries and is highly sensitive to linguistic variation compared to rigid reasoning-chain formats~\cite{openai2024learning}. Besides, prior work often overlooks leveraging the detailed reasoning traces from powerful reasoning LLMs~\cite{guo2025deepseek}, despite their rich intermediate signals that can effectively guide student learning.

\begin{figure*}[htb]
    \centering
     \includegraphics[width=1\textwidth,height=0.45\textwidth]{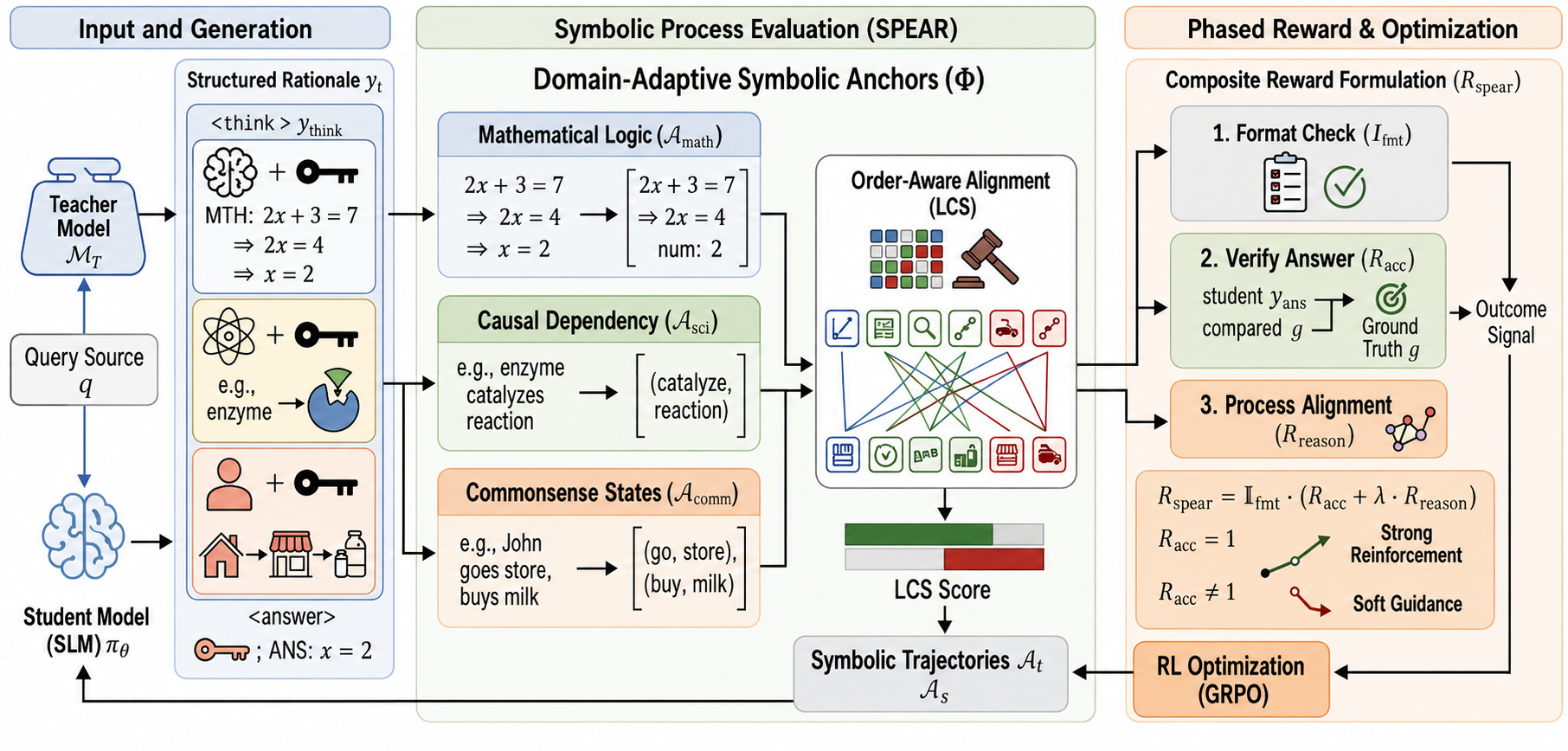}
     \caption{Overview of \textbf{SPEAR} (\textbf{S}ymbolic \textbf{P}rocess \textbf{E}valuation and \textbf{A}lignment \textbf{R}eward) method. It supervises the "logical skeleton" of reasoning tasks by aligning symbolic milestones extracted from teacher and student traces. By using an order-aware alignment (LCS) and an outcome-process combined reward, the framework enables small language models to internalize reasoning processes rather than token imitation from teacher's responses.}
    \label{fig_overview}
    \vspace{-0.5cm}
\end{figure*}

To bridge these gaps, we propose \textbf{SPEAR} (\textbf{S}ymbolic \textbf{P}rocess \textbf{E}valuation and \textbf{A}lignment \textbf{R}eward), a training-free, plug-and-play process reward framework for sequence-level on-policy distillation, as illustrated in Figure~\ref{fig_overview}. Instead of forcing token-level matching or relying on heavy neural PRMs, SPEAR distills the logical structure of the teacher's rationale. We project high-dimensional reasoning traces into domain-adaptive \textit{symbolic trajectories}—such as computational processes for math, causal dependencies for science, and entity state-transitions for commonsense. By employing the longest common subsequence (LCS) to align the student's explored symbolic path with the teacher's reference milestones, SPEAR provides a dense and order-aware reward. This enforces chronological logical consistency while granting SLMs the freedom to formulate their own response.

In summary, the contributions of our work are:
\vspace{-5pt}
\begin{enumerate}
    \item We introduce \textbf{SPEAR}, a novel, training-free process reward framework that enables highly efficient sequence-level on-policy distillation without the computational burden of process reward models.
    \item SPEAR uses teacher thinking processes containing rich reasoning signals and projects these high-dimensional traces into \textbf{domain-adaptive symbolic anchors}, providing the student with a structural reference that successfully extends process supervision to reasoning tasks.
    \item We conduct experiments across \textbf{comprehensive reasoning benchmarks}, including math, science, and commonsense tasks. Experimental results demonstrate that integrating SPEAR into standard RL pipelines consistently outperforms baselines, validating the efficacy of decoupling logical acquisition from linguistic expression.
\end{enumerate}

\section{Related Work}
\textbf{Reinforcement Learning for Reasoning} \ \ Reinforcement learning with verifiable rewards (RLVR) has become the primary driver for inducing multi-step reasoning behaviors in LLMs~\cite{luong2024reft, guo2025deepseek}. Algorithms like Group Relative Policy Optimization (GRPO)~\cite{shao2024deepseekmath} and its subsequent refinements—e.g., Dr. GRPO~\cite{liu2025understanding}, which analyzes and mitigates length-related optimization bias in relative policy updates, and Dynamic Sampling Policy Optimization (DAPO)~\cite{yu2025dapo}, which introduces dynamic trajectory reweighting to reduce sampling and length-induced biases, have substantially improved the efficiency and stability of RL-based reasoning training.  However, these methods typically rely on sparse, binary outcome rewards~\cite{zhang2026liralocalcrosslayerinformation}.  This sparsity can lead to "overthinking," where models generate thousands of redundant tokens to maximize rewards without valid logical progress~\cite{agarwal2024policy,song2025walk}. \\
\textbf{LLMs Distillation} \ \ Distillation methods differ in trajectory source and teacher supervision. Off-policy distillation commonly uses teacher-generated responses, whereas on-policy distillation learns from student-generated trajectories~\cite{wen2025reasoning,song2026survey}. Teacher guidance may use token-distribution matching or teacher-derived rewards and can be combined with task-reward optimization~\cite{xu2025kdrl}. RLKD~\cite{xu2025distilling} uses structural alignment, while on-policy self-distillation~\cite{zhao2026self} uses privileged information. Skill-SD~\cite{wang2026skillsd} combines RL with an auxiliary token-level self-distillation loss. SPEAR instead incorporates fixed teacher rationales directly into a symbolic alignment reward for student rollouts, without teacher token probabilities. \\
\textbf{Process Supervision and Symbolic Alignment} \ \ Evaluating intermediate reasoning steps is critical for mitigating reward hacking and reducing sample complexity~\cite{lightman2024let, song2025prmbench, li2024process}. While neural PRMs offer dense supervision, they are computationally expensive and prone to hallucinations in domain-specific tasks~\cite{yu2024reasoning, su2025reinforcement}. Rule-based process rewards, such as RePAIR~\cite{wangrepair} and Logic-RL~\cite{xie2025logic}, provide a verifiable, training-free alternative. Specifically, the use of longest common subsequence (LCS) rewards has shown promise in identifying optimal reasoning paths across multiple tasks~\cite{dong2025enhancing}. SPEAR extends this by mapping diverse reasoning traces into symbolic, order-aware trajectories that provide a general alignment signal.

\section{Method}
Our approach departs from traditional token-level distillation by framing the reasoning process as a sequence of symbolic milestones. By supervising the student model through alignment with these milestones, SPEAR could internalize the underlying logical structure of a task while retaining the flexibility to explore their own linguistic paths.

\subsection{Teacher-Guided Policy Optimization}
For a query $q$, teacher $\mathcal{M}_T$ provides a structured response $y^T=(y_{think},y_{ans})$ offline. Its rationale remains a fixed reward reference. During RL, student responses are sampled from the current rollout policy $\pi_{\theta_{\mathrm{old}}}$ and scored against that reference. Here, ``on-policy'' describes student rollout generation, not when teacher references are collected. Teacher knowledge enters through symbolic alignment rewards rather than a teacher--student token-distribution loss. We optimize the student policy $\pi_\theta$ using RL algorithms such as Group Relative Policy Optimization (GRPO)~\cite{guo2025deepseek}, with the following clipped surrogate objective:

\begin{equation}
\resizebox{0.89\columnwidth}{!}{%
$\mathcal{J}(\theta)
=
\mathbb{E}_{q \sim \mathcal{D},\, o \sim \pi_{\theta_{\mathrm{old}}}(\cdot \mid q)}
\left[
\min\Big(
r(o)\hat{A},
\operatorname{clip}(r(o), 1-\epsilon, 1+\epsilon)\hat{A}
\Big)
\right]$
}
\end{equation}
where $r(o)=\frac{\pi_{\theta}(o \mid q)}{\pi_{\theta_{\mathrm{old}}}(o \mid q)}$, and the advantage $\hat{A}$ is calculated from our composite SPEAR reward. This objective encourages the student to explore diverse reasoning paths and optimize for correctness and logical consistency, rather than merely reproducing the teacher's surface form.

\subsection{Domain-Adaptive Symbolic Anchors}
\label{subsec:anchors}
The core of SPEAR is the projection function $\Phi: \mathcal{Y} \rightarrow \mathcal{A}$, which maps a high-dimensional natural language trace $y$ into a low-dimensional \textit{Symbolic Trajectory} $\mathcal{A}_t = [a_1, a_2, \dots, a_n], \quad a_i \in \mathcal{V}_t$. We define domain-specific anchors $a_i$ grounded in symbolic logic and linguistic dependency theory.

\subsubsection{Formal Logic and Mathematical States}
In quantitative reasoning, the structure is defined by the evolution of symbolic expressions and variable assignments. Inspired by symbolic state-space search \cite{lample2019deep}, we define the mathematical anchor extraction function $\Phi_{\mathrm{math}}$ as a regex-based projection that isolates the calculus of the reasoning trace from its natural language prose. 

\begin{equation}
\mathcal{V}_{\mathrm{math}} := \mathcal{V}_{\mathrm{latex}} \cup \mathcal{V}_{\mathrm{assign}}
\end{equation}
where $\mathcal{V}_{\mathrm{latex}}$ captures expressions wrapped in LaTeX 
delimiters (\texttt{\$...\$}, \texttt{\textbackslash[...\textbackslash]}) and 
$\mathcal{V}_{\mathrm{assign}}$ captures explicit assignments (e.g., \textit{``$x=5$''}). Notably, we exclude bare numerical values from the extraction to ensure that the alignment reward measures the \textbf{structural derivation} of the problem rather than the occurrence of constants. By focusing on these symbolic transitions, $\Phi_{\mathrm{math}}$ captures the trajectory of the problem-solving process, ensuring the student model respects the necessary order of operations, such as the isolation of variables prior to substitution.


\subsubsection{Causal Dependency Anchors for Science}

For scientific reasoning, logic is embedded in interactions between domain-specific entities and actions. Standard overlap metrics fail here because they ignore the directionality of these relations. Grounded in the finding that syntactic structure is a primary driver of semantic role representation~\cite{punyakanok2008importance}, we implement $\Phi_{\mathrm{sci}}$ using a dependency-parsing framework following successful applications in large-scale scientific relation extraction~\cite{percha2018global}. This yields a de-duplicated relational sequence where each anchor $a_i$ is a verb-centered relational tuple, with $v$, $sbj$, $obj$ denoting the governing verb, subject, and object respectively.

\begin{equation}
\resizebox{0.95\columnwidth}{!}{$
\mathcal{V}_{\mathrm{sci}} :=
\begin{cases}
(\mathrm{lemma}(v), \mathrm{span}(obj)) & \text{if object } obj \text{ exists} \\
(\mathrm{span}(sbj), \mathrm{lemma}(v)) & \text{otherwise}
\end{cases}
$}
\end{equation}

To prevent reward inflation through repetitive reasoning loops, $\mathcal{A}_{\mathrm{sci}}$ enforces a uniqueness constraint: $\forall\, i \neq j: a_i \neq a_j$, ensuring SPEAR rewards progression through the teacher's causal chain rather than dwelling on a single repeated fact.

\subsubsection{Open-Domain State Transitions}
In more general open-domain reasoning, such as commonsense and open-ended tasks, the logic typically tracks the movement of agents or the evolution of objects through temporal states. Grounded in \textit{Script Theory}~\cite{schank2013scripts}, which posits that knowledge is organized around stereotypical event sequences, we implement $\Phi_{\mathrm{com}}$ as a projection into a sequence of State-Action anchors. We extract these anchors by identifying noun phrases (noun chunks) and their governing syntactic heads. Formally, for each noun chunk $c$, $\mathcal{V}_{\mathrm{com}}$ is defined as:

\begin{equation}
\resizebox{0.95\columnwidth}{!}{$
\mathcal{V}_{\mathrm{com}} :=
\begin{cases}
(\mathrm{root}(c),\ \mathrm{lemma}(\mathrm{head}(c))) & \text{if } \mathrm{head}(c) \in \mathrm{VERB} \\
\mathrm{root}(c) & \text{otherwise}
\end{cases}
$}
\end{equation}
These anchors act as logical key-frames, capturing the student’s ability to track world-state changes chronologically (e.g., \textit{ball, throw''} $\rightarrow$ \textit{window, break''}). By extracting the \text{root} of the noun phrase and the \text{lemma} of the action, it ensures that the alignment reward is invariant to modifiers and tense variations, focusing purely on the \textbf{state transition} intended by the teacher.

\subsection{Sequential Alignment via LCS-F1}
Motivated by the success of longest common subsequence (LCS) rewards in identifying coherent reasoning trajectories across diverse tasks~\cite{dong2025enhancing}, we employ LCS to align chronological logical consistency between the student trajectory $\mathcal{A}_s$ and the teacher trajectory $\mathcal{A}_t$. Unlike unordered metrics such as Jaccard similarity or surface-level metrics like ROUGE~\cite{lin2004rouge}, LCS is strictly \textbf{order-aware} and invariant to redundant linguistic filler that does not contribute to the symbolic path.

We define the reasoning process reward as the \textbf{LCS-F1} score:
\begin{equation}
    \label{eq:lcs_f1}
    R_{reason} = \frac{2|\text{LCS}(\mathcal{A}_s, \mathcal{A}_t)|}{|\mathcal{A}_s| + |\mathcal{A}_t|},
\end{equation}

which is the harmonic mean of alignment precision $|\mathrm{LCS}|/|\mathcal{A}_s|$ and teacher-milestone recall $|\mathrm{LCS}|/|\mathcal{A}_t|$. Precision penalizes spurious or repetitive student anchors, preventing verbose trajectories from inflating the reward, while recall penalizes trajectories that omit teacher milestones. Consequently, a short matching prefix cannot receive full credit, and the maximum score is attained only when the extracted trajectories align completely. This balance encourages both teacher-milestone coverage and high \textbf{logical density}. More analysis of LCS-F1 is provided in Appendix~\ref{sec:appendix_lcs_norm}.

To ensure the reward is computable efficiently during real-time RL rollouts, we use a dynamic programming approach where the LCS alignment state $L(i, j)$ is computed as follows:
\begin{equation}
\resizebox{1.0\columnwidth}{!}{%
    $L(i, j) = \begin{cases} 
      L(i-1, j-1) + 1 & \text{if } a_{s,i} = a_{t,j} \\
      \max(L(i, j-1), L(i-1, j)) & \text{otherwise}
   \end{cases}$
}
\end{equation}
This provides a dense reward signal that penalizes logical reversals while rewarding the correct chronological order of reasoning milestones.

\subsection{Gated Composite Reward Formulation}
To stabilize the RL process, we integrate SPEAR into a phased reward structure~\cite{xie2025logic}. We implement a strict gate for formatting compliance $R_{fmt}$, checking \texttt{<think>} and \texttt{<answer>} blocks. It follows by a weighted combination of accuracy $R_{acc}$ and reasoning process alignment $R_{reason}$. The total reward $R_{spear}$ is defined as:
\begin{equation}
\label{eq:reward}
\begin{aligned}
    R_{spear} &= \mathbb{I}_{\mathrm{fmt}} \cdot (R_{acc} + \lambda \cdot R_{reason}) \\
\end{aligned}
\end{equation}
where $\mathbb{I}_{\mathrm{fmt}} \in \{0, 1\}$ is an indicator function for format compliance. In our implementation, we set $\lambda = 0.5$. Unlike purely outcome-based supervision, which provides a sparse binary signal (0 or 1), $R_{reason}$ serves as a \textbf{dense shaping signal} in $[0, 1]$. By awarding \textbf{partial credit} for reasoning milestones even when the final answer is incorrect, this design prevents gradient collapse during early training phases and encourages the model to refine its logical trajectory throughout the exploration.

\begin{algorithm}[h]
\caption{SPEAR Reward Calculation}
\label{alg:spear}
\begin{algorithmic}[1]
\Procedure{ComputeSPEAR}{$y_s, y_t, y_{gold}, \tau$}
    \Comment{$y_s$: student response, $y_t$: teacher response, $y_{gold}$: gold answer, $\tau$: task type}
\State $y_{think}, y_{ans} \gets \text{ParseTags}(y_s)$
\If{$\neg \text{ValidFormat}(y_s)$}
    \State \Return $0.0$
\EndIf
\State $R_{acc} \gets \text{VerifyAnswer}(y_{ans}, y_{gold})$
\State $\mathcal{A}_s \gets \Phi(y_{think}, \tau)$
\State $\mathcal{A}_t \gets \Phi(y_t, \tau)$
\If{$\mathcal{A}_s = \emptyset \lor \mathcal{A}_t = \emptyset$}
    \State \Return $R_{acc}$
\EndIf
\State $\ell \gets \text{LCS}(\mathcal{A}_s, \mathcal{A}_t)$
\State $R_{reason} \gets 2\ell / (|\mathcal{A}_s| + |\mathcal{A}_t|)$ \Comment{LCS-F1}
\State \Return $R_{acc} + \lambda \cdot R_{reason}$
\EndProcedure
\end{algorithmic}
\end{algorithm}

\section{Experiments}
\subsection{Datasets}
We focus on evaluating reasoning abilities with various datasets, including mathematical reasoning with GSM8K~\cite{cobbe2021training} and MATH~\cite{hendrycks2021measuring}, scientific reasoning with GPQA~\cite{rein2024gpqa}, and commonsense reasoning with CommonsenseQA~\cite{talmor2019commonsenseqa}. Datasets statistics are shown in Appendix~\ref{sec:appendix_datasets}.

\begin{table*}[ht]
\centering
\small 
\begin{tabular}{l|cc|cc|c}
\toprule
\multirow{2}{*}{\textbf{Method}} & \multicolumn{2}{c|}{\textbf{Mathematical}} & \textbf{Scientific} & \textbf{Commonsense} & \multirow{2}{*}{\textbf{Avg. $\Delta$}} \\
 & GSM8K & MATH & GPQA & CommonsenseQA & \\ 
\midrule
\rowcolor{gray!15} \textbf{Teacher LLM} & & & & & \\
DeepSeek-V3.2 & 95.75 & 90.20 & 59.10 & 90.16 & \\ 
\midrule
\rowcolor{gray!15} \textbf{Llama-3-8B-Instruct} & & & & & \\
Zero-shot (Base) & 67.10 & 26.80 & 25.75 & 75.67 &  \\
SFT (Distilled) & 73.54 & 30.40 & 33.83 & 74.75 &  \\ \hline
GRPO & 78.54 & 28.20 & 29.29 & 75.74 & -- \\
GRPO + Logic-RL \cite{xie2025logic} & 79.00 & 29.00 & 29.80 & 74.59 & +0.16\% \\
\textbf{GRPO + SPEAR (Ours)} & \textbf{81.05} & \textbf{30.60} & \textbf{31.82} & \textbf{76.97} & \textbf{+2.17\%} \\ \hline
Dr. GRPO & 75.28 & 30.00 & 32.83 & 77.05 & -- \\
Dr. GRPO + Logic-RL \cite{xie2025logic} & 75.97 & 31.20 & 32.32 & 77.46 & +0.45\% \\
\textbf{Dr. GRPO + SPEAR (Ours)} & \textbf{77.71} & \textbf{32.80} & \textbf{34.34} & \textbf{78.28} & \textbf{+1.99\%} \\ \hline
DAPO & 77.79 & 31.60 & 30.81 & 77.54 & -- \\
DAPO + Logic-RL \cite{xie2025logic} & 78.24 & 32.40 & 31.31 & 77.05 & +0.32\% \\
\textbf{DAPO + SPEAR (Ours)} & \textbf{80.44} & \textbf{33.60} & \textbf{32.83} & \textbf{78.61} & \textbf{+1.94\%} \\ 
\midrule
\rowcolor{gray!15} \textbf{Qwen3-4B} & & & & & \\
Zero-shot (Base) & 88.10 & 67.40 & 35.35 & 83.95 &  \\
SFT (Distilled) & 90.98 & 70.20 & 41.41 & 80.90 &  \\ \hline
GRPO & 92.11 & 71.40 & 37.37 & 86.98 & -- \\
GRPO + Logic-RL \cite{xie2025logic} & 92.49 & 71.80 & 37.88 & 87.05 & +0.34\% \\
\textbf{GRPO + SPEAR (Ours)} & \textbf{93.93} & \textbf{73.80} & \textbf{40.91} & \textbf{88.36} & \textbf{+2.29\%} \\ \hline
Dr. GRPO & 91.50 & 72.00 & 39.40 & 85.01 & -- \\
Dr. GRPO + Logic-RL \cite{xie2025logic} & 91.81 & 72.60 & 39.90 & 85.08 & +0.37\% \\
\textbf{Dr. GRPO + SPEAR (Ours)} & \textbf{93.03} & \textbf{74.40} & \textbf{41.92} & \textbf{86.39} & \textbf{+1.96\%} \\ \hline
DAPO & 92.49 & 72.80 & 40.40 & 86.00 & -- \\
DAPO + Logic-RL \cite{xie2025logic} & 92.87 & 73.40 & 40.91 & 85.74 & +0.31\% \\
\textbf{DAPO + SPEAR (Ours)} & \textbf{94.54} & \textbf{74.80} & \textbf{42.93} & \textbf{87.21} & \textbf{+1.95\%} \\
\bottomrule
\end{tabular}
\caption{\label{table_results} Accuracy (\%) across various reasoning tasks with different distillation methods. Results demonstrate that SPEAR consistently outperforms other outcome-only baselines across multiple RL frameworks and student architectures. Avg. $\Delta$ denotes the average improvement over the corresponding base RL method. Best results for each student model are \textbf{bolded}.}
\end{table*}

\subsection{Baselines}
We compare against \textbf{SFT distillation} and the following \textbf{RL baselines}, whose rewards use answer correctness and formatting without teacher-rationale guidance; SPEAR adds that guidance through symbolic alignment:
\begin{itemize}
    \item \textbf{GRPO}~\cite{guo2025deepseek}: The standard Group Relative Policy Optimization optimized via outcome-based rewards.
    \item \textbf{Dr. GRPO}~\cite{liu2025understanding}: A bias-corrected variant of GRPO optimized with outcome-based signals.
    \item \textbf{DAPO}~\cite{yu2025dapo}: A decoupled clip and dynamic sampling policy optimization method.
\end{itemize}
We also include rule-based reward baseline \textbf{Logic-RL}~\cite{xie2025logic}, which gives partial reward when the format or final answer is not fully correct.

\subsection{Models}
Since we aim to distill the teacher's thinking process, we employ DeepSeek-V3.2~\citep{deepseek2025v32} as the teacher LLM, owing to its strong thinking mode and relatively low cost. For student SLMs, we choose Llama-3-8B-Instruct~\citep{dubey2024llama} for its original reasonable performance on the chosen benchmarks, as well as Qwen3-4B~\cite{yang2025qwen3} to test the generalizability of the SPEAR method.  \\
All evaluation results are based on the zero-shot and the average of three independent runs. We use spaCy~\citep{honnibal2020spacy} and \texttt{en\_core\_web\_sm} as tools to extract dependency anchors. We set $\lambda=0.5$ in Equation~\ref{eq:reward} for the main experiments. More implementation details are in Appendix~\ref{sec:appendix_implement}.

\subsection{Main Results}
Main results are shown in Table~\ref{table_results}. \\
\textbf{Insights of Distillation} \ \ Across both student architectures, SFT and RL generally improve base performance, but the effect is \textbf{task-dependent}. The gains are most consistent on mathematical reasoning, where both SFT and RL improve GSM8K and MATH. However, the improvement is smaller when the base model is already strong or near \textbf{saturation}. This is especially evident for Qwen3-4B on GSM8K (88.10\%), leaving limited room for further improvement. Distillation is also not uniformly beneficial for commonsense reasoning: for example, Qwen3-4B drops from 83.95\% to 80.90\%, and Llama3-8B-Instruct drops from 75.67\% to 74.75\% on CommonsenseQA under SFT, suggesting that forcing the student to directly learn from teacher's natural language, which may contain implicit and less transferable patterns, can impair the distillation. \\
\textbf{Comparison of Supervised Distillation and Outcome-Based RL} \ \ The table reveals a clear trade-off between supervised distillation and outcome-based RL. On the mathematical benchmarks, RL is generally more effective than SFT, consistent with the fact that the \textbf{final answer is strongly coupled with stepwise reasoning in math problems}. By contrast, SFT remains highly competitive on GPQA, especially for Llama-3-8B-Instruct, where SFT achieves 33.83\%, exceeding all three plain RL baselines. A similar pattern is also observed for Qwen3-4B. This suggests that for multi-choice scientific tasks such as GPQA, it needs complex and scientific reasoning processes, while it's still possible for models to  \textbf{guess the final option correctly}. Thus, direct imitation of teacher explanations can sometimes transfer more useful domain knowledge than outcome-only exploration. \\
\textbf{SPEAR Advantages over Baselines} \ \ For all tasks, our SPEAR method consistently improves over the baselines across different SLMs and outcome-based RL baselines. These gains are more evident on scientific and mathematical reasoning tasks, where the extracted anchors capture explicit causal and computational structures. In contrast, CommonsenseQA shows the smallest improvement, suggesting that \textbf{commonsense reasoning is harder to analyze and reward using extracted anchors}, because it relies more on natural language variation and implicit structural patterns. Moreover, SPEAR consistently outperforms the reward baseline Logic-RL, indicating that format and final answer-based \textbf{partial reward signals are not sufficient for reliable distillation}. In fact, such weak supervision can even hurt performance when the rationale is loosely correlated with the final answer, as shown in the case of scientific and commonsense tasks.

\section{Discussion}
\subsection{Performance-Efficiency Analysis}
To study the tradeoff between SPEAR and process reward models (PRMs), we compare against Qwen2.5-Math-PRM-7B~\cite{zhang2025lessons}, a math-specialized PRM, and VersaPRM~\cite{zeng2025versaprm}, a multi-domain PRM. We run the same distillation setup on Qwen3-4B and only replace the reward signal with the average process reward produced by each PRM. The results are reported in Table~\ref{table_prms}. More resource overhead comparisons are provided in Appendix~\ref{sec:appendix_efficiency}.

Qwen2.5-Math-PRM-7B consistently achieves the best performance on MATH, which is expected given its domain-specific supervision. However, its advantage over SPEAR is only 0.32 percentage points on average across the three RL frameworks. In contrast, it transfers poorly to GPQA, where it underperforms SPEAR by an average of 3.37 percentage points. This suggests that a domain-specific PRM can have limited transferability to tasks outside its target domain.

VersaPRM, as a multi-domain PRM, provides substantially stronger GPQA supervision than the math-only PRM, improving accuracy by 2.86 percentage points on average. Relative to VersaPRM, SPEAR is 0.26 points higher on MATH and 0.51 points higher on GPQA on average across the three RL frameworks, although VersaPRM leads in two individual settings. Importantly, SPEAR relies only on lightweight processing with spaCy and LCS, where the only external component is the small \texttt{en\_core\_web\_sm} model ($\sim$12MB). This stands in sharp contrast to neural PRMs, which require an additional billion-parameter model for process supervision. Overall, the result demonstrates that SPEAR provides a highly \textbf{favorable performance-efficiency tradeoff}, delivering competitive performance with significantly lower computational cost.

\begin{table}[htb]
\centering
\resizebox{1.0\columnwidth}{!}{\begin{tabular}{l|cc|c}
\toprule
\textbf{Qwen3-4B} & \textbf{MATH} & \textbf{GPQA} & \textbf{Overhead}\\
\midrule
GRPO + Qwen2.5-Math-PRM & \textbf{74.30} & 37.37 & 7B Neural PRM \\
GRPO + VersaPRM & 73.69 & \textbf{41.41} & 8B Neural PRM \\
GRPO + SPEAR & 73.80 & 40.91 & spaCy + LCS ($\sim$12MB) \\
\midrule
Dr. GRPO + Qwen2.5-Math-PRM & \textbf{74.45} & 38.89 & 7B Neural PRM \\
Dr. GRPO + VersaPRM & 73.61 & 40.40 & 8B Neural PRM \\
Dr. GRPO + SPEAR & 74.40 & \textbf{41.92} & spaCy + LCS ($\sim$12MB) \\
\midrule
DAPO + Qwen2.5-Math-PRM & \textbf{75.21} & 39.40 & 7B Neural PRM \\
DAPO + VersaPRM & 74.91 & 42.43 & 8B Neural PRM \\
DAPO + SPEAR & 74.80 & \textbf{42.93} & spaCy + LCS ($\sim$12MB) \\
\bottomrule
\end{tabular}}
\caption{\label{table_prms}
Performance–efficiency tradeoff between neural PRMs and SPEAR on Qwen3-4B. SPEAR achieves competitive performance while using only a lightweight \texttt{en\_core\_web\_sm} ($\sim$12MB) parser, compared to billion-parameter neural PRMs.
}
\end{table}

\subsection{Ablation of Reward Functions}
Table~\ref{table_ablation} analyzes the contribution of each component in SPEAR. \\
\textbf{Reward Components} Removing the format gate, which enforces the \texttt{<think>} and \texttt{<answer>} structure, leads to a consistent drop on both benchmarks, indicating that structured outputs help stabilize training and enable reliable reward computation. In contrast, removing the accuracy reward results in the largest degradation ($30.60 \rightarrow 25.40$ on MATH and $31.82 \rightarrow 26.77$ on GPQA), confirming that final-answer supervision remains the dominant learning signal in RL-based distillation. \\
\textbf{Reasoning Weight} We vary the coefficient $\lambda$ in Equation~\ref{eq:reward}, where $\lambda=0.5$ is used in our main experiments. Setting $\lambda=0$ removes the process reasoning reward and reduces performance on both tasks, showing that outcome-only supervision is insufficient. MATH performs best with $\lambda=0.5$, while GPQA peaks at a higher value ($\lambda=0.75$). However, further increasing $\lambda$ to $1.0$ degrades performance, suggesting that over-emphasizing reasoning alignment can harm optimization, and $\lambda$ needs to be optimized for different tasks. \\
\textbf{Alignment Design} Replacing symbolic anchors with raw-token LCS-F1 reduces average accuracy by 2.37 percentage points, while removing order information with Jaccard overlap reduces it by 2.21 points. This indicates that SPEAR benefits from both abstraction (symbolic anchors) and sequential structure (order-aware alignment), which together better capture the causal nature of reasoning processes. \\
\textbf{Anchor Extraction} Using generic surface-level anchors (e.g., logic markers such as ``however'', ``wait'', ``thus'') instead of our task-specific extraction reduces accuracy by 2.77 percentage points on average. This suggests that domain-aware milestone extraction is crucial for providing informative reasoning signals. 

Overall, these results validate that SPEAR improves distillation through the joint effect of format constraints, answer supervision, and order-aware symbolic reasoning alignment, with each component playing a distinct and necessary role.

\begin{table}[htb]
\centering
\small
\resizebox{1.0\columnwidth}{!}{\begin{tabular}{l|cc}
\toprule
\textbf{Llama-3-8B-Instruct} & \textbf{MATH} & \textbf{GPQA} \\
\midrule
SPEAR (GRPO, $\lambda = 0.5$) & \textbf{30.60} & 31.82 \\
\midrule
\rowcolor{gray!10} \textit{Reward Components} & & \\
\quad w/o Format Gate & 30.00 & 30.30 \\
\quad w/o Accuracy Reward & 25.40 & 26.77 \\
\midrule
\rowcolor{gray!10} \textit{Reasoning Weight ($\lambda$)} & & \\
\quad $\lambda = 0.0$ (w/o process reasoning) & 28.40 & 29.80 \\
\quad $\lambda = 0.25$ & 29.40 & 30.30 \\
\quad $\lambda = 0.75$ & 29.00 & \textbf{32.32} \\
\quad $\lambda = 1.0$ & 28.40 & 30.81 \\
\midrule
\rowcolor{gray!10} \textit{Alignment Design} & & \\
\quad Raw Token LCS-F1 & 28.40 & 29.29 \\
\quad Unordered Symbol Overlap (Jaccard) & 28.20 & 29.80 \\
\midrule
\rowcolor{gray!10} \textit{Anchor Extraction} & & \\
\quad Generic Surface Anchors & 28.60 & 28.28 \\
\bottomrule
\end{tabular}}
\caption{\label{table_ablation} Ablation of SPEAR reward design on Llama-3-8B-Instruct. Results highlight the importance of accuracy supervision, balanced process reasoning weight, and order-aware symbolic alignment.}
\end{table}

\subsection{Effect of Cold-Start SFT on RL Distillation}
To further study the interaction between cold-start SFT and SPEAR-based RL in distillation, we vary the fraction of data used for SFT before RL and evaluate how the final performance changes. Specifically, we train Qwen3-4B on the MATH dataset using teacher-generated responses. We cold-start the model via SFT on progressively larger fractions of the training data (10\% increments), and subsequently apply GRPO with SPEAR on the rest of the training data to evaluate performance improvements.

\begin{figure}[h]
    \centering
     \includegraphics[width=0.49\textwidth]{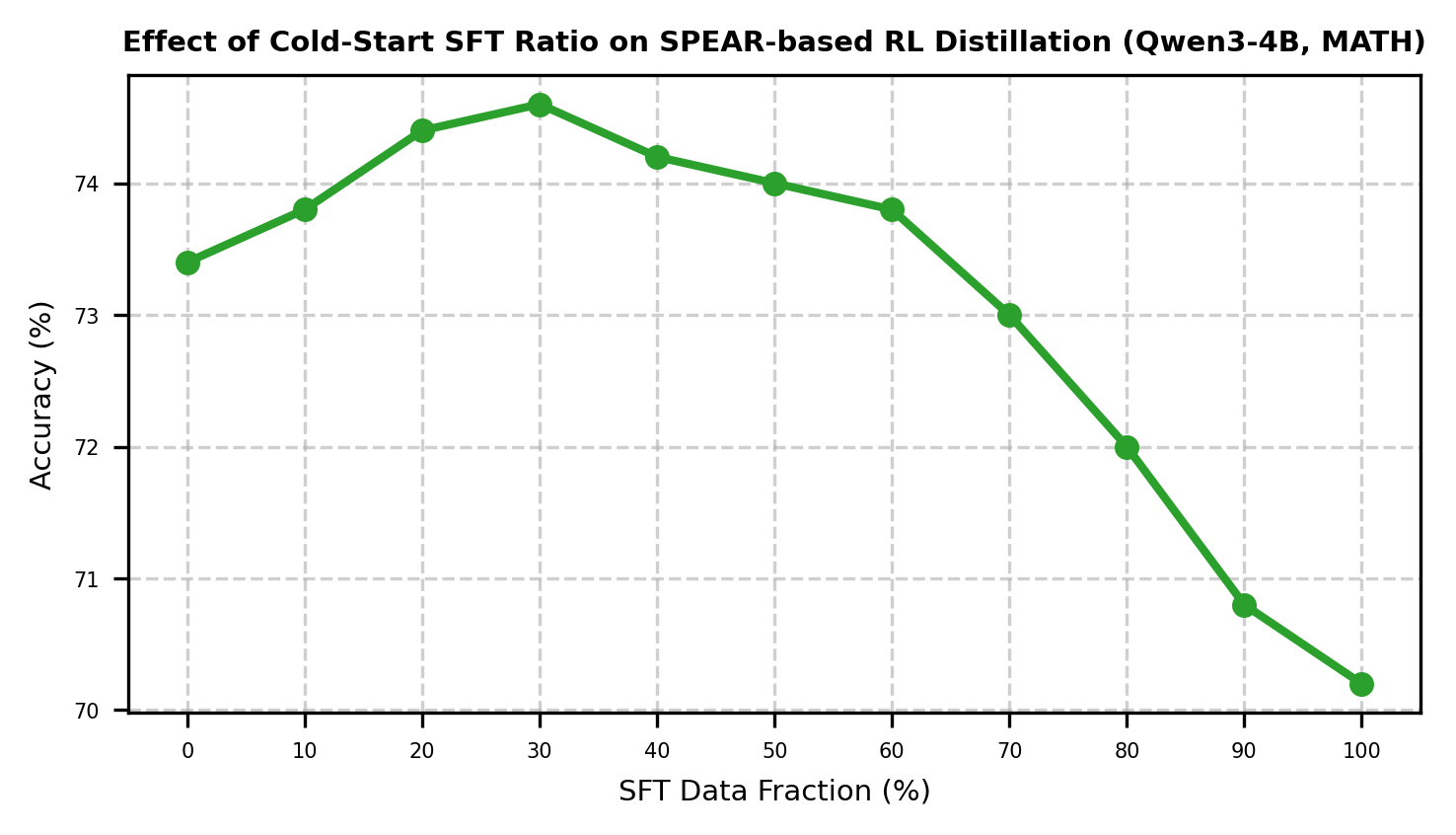}
     \caption{Effect of cold-start SFT ratio on SPEAR-based RL distillation (Qwen3-4B, MATH). A moderate amount of SFT data yields the best result, while excessive SFT reduces the subsequent RL performance.}
    \label{fig_sft-rl}
    \vspace{-0.45cm}   
\end{figure}

The result is shown in Figure~\ref{fig_sft-rl}. Starting from no SFT (100\% SPEAR-based RL), introducing a small amount of supervised data consistently improves performance, as SFT helps the model learn basic reasoning patterns and output structure before RL optimization. The best performance is achieved when using a moderate fraction of SFT data (30\%), suggesting that SFT provides a strong initialization that enhances the effectiveness of subsequent RL training. However, as the proportion of SFT data increases further, performance gradually declines. This is likely because excessive SFT reduces the diversity of exploration during RL, causing the model to overfit to teacher-generated responses. Overall, the results demonstrate that a balanced combination of SFT and SPEAR-based RL is crucial, where SFT provides a stable starting point, and RL refines reasoning through reward-driven optimization.

\subsection{Qualitative Case Study}
To illustrate how SPEAR transfers reasoning structure, we present a representative GPQA example and compare the anchors extracted from the teacher and student rationales. For GPQA, our extractor forms anchors as verb-object pairs via dependency parsing, so the comparison focuses on whether the student preserves the teacher's ordered causal relations in its generated \texttt{<think>} trace.

\begin{tcolorbox}[
    float, floatplacement=h!,
    colback=gray!5, colframe=black!70,
    title=\textbf{Case Study: SPEAR Reward Signal on GPQA},
    fonttitle=\small, boxrule=0.5pt, left=4pt, right=4pt]
\small
\textbf{Question:} Trans-cinnamaldehyde was treated with methylmagnesium bromide (product 1); then with pyridinium chlorochromate (product 2); then with (dimethyl(oxo)-l6-sulfaneylidene)methane in DMSO at elevated temperature (product 3). How many carbon atoms are in product 3? \\
\textbf{Teacher Response} $(\mathcal{T})$: \texttt{<think>} Methylmagnesium bromide \underline{adds} a \underline{methyl group} to the aldehyde, giving a 10-carbon alcohol. Then, pyridinium chlorochromate \underline{oxidizes} the \underline{alcohol} to a ketone. Dimethyloxosulfonium methylide \underline{adds} one \underline{carbon unit} to form a cyclopropyl ring. \texttt{</think>} \texttt{<answer>} Thus, the product contains 11 carbon atoms. \texttt{</answer>} \\
\textbf{Student Response} $(\mathcal{S})$: \texttt{<think>} The Grignard reagent \underline{adds} a \underline{methyl group} to the aldehyde. Pyridinium chlorochromate \underline{oxidizes} the \underline{alcohol} to a ketone. The sulfonium ylide \underline{removes} a \underline{methylene unit} to the alkene. \texttt{</think>} \texttt{<answer>} 11 \texttt{</answer>} \\[2pt]

The underlined anchors are compared in sequence :
\smallskip
\resizebox{1.0\columnwidth}{!}{%
\begin{tabular}{lll}
\toprule
$\mathcal{T}$ Anchors & $\mathcal{S}$ Anchors & Match \\
\midrule
\texttt{add methyl group} & \texttt{add methyl group} & \textcolor{teal}{\checkmark} \\
\texttt{oxidize alcohol}  & \texttt{oxidize alcohol}  & \textcolor{teal}{\checkmark} \\
\texttt{add carbon unit}  & \texttt{remove methylene unit} & \textcolor{orange}{$\times$} \\
\bottomrule
\end{tabular}}
\smallskip

$\text{LCS}(\mathcal{T}, \mathcal{S}) = 2$, \quad
$R_{\text{reason}} = \frac{2|\text{LCS}|}{|\mathcal{T}|+|\mathcal{S}|} = \frac{4}{6} = 0.67$
\end{tcolorbox}

Notably, verb-object anchors correctly absorb the subject paraphrase in step 1 and 2, yielding an exact match regardless of how the reagent is named. Step 3 fails (\texttt{remove methylene unit} vs.\ \texttt{add carbon unit}) due to lexical divergence in the object. Crucially, \textbf{both responses reach the correct answer}, so the accuracy reward assigns identical scores to both; SPEAR's combined reward penalizes the imprecise terminology in step 3, providing a finer training signal that accuracy alone cannot.

\section{Conclusion}
In this work, we introduce SPEAR, a training-free reward framework for sequence-level on-policy distillation. Across math, science, and commonsense tasks, SPEAR provides dense process supervision with a favorable efficiency–performance tradeoff. Our ablations also show that balanced reasoning weight, order-aware symbolic alignment, and a moderate SFT cold start are all important for effective distillation. We hope this work inspires efficient process supervision in distillation.

\section*{Limitations}
Despite its effectiveness, SPEAR has several limitations. First, considering of the computation overhead, our method relies on symbolic anchor extraction and exact subsequence matching, which may fail to fully capture semantically equivalent but structurally different reasoning trajectories. Consequently, valid alternative reasoning paths that diverge from the teacher’s symbolic ordering can receive lower rewards, particularly in open-ended reasoning tasks. Second, SPEAR assumes that teacher-generated rationales provide reliable reasoning trajectories, although LLM explanations may contain redundant or suboptimal intermediate steps. Additionally, our current implementation depends on rule-based extraction pipelines, such as regex matching and dependency parsing, which may be sensitive to parsing errors and formatting inconsistencies. Finally, due to computational constraints, we focus exclusively on sequence-level on-policy distillation and do not explore token-level reverse-KL distillation, which remain promising directions for future work.


\section*{Ethics Statement}
\textbf{Use of AI Assistants} \ \ We used AI assistants for manuscript editing, coding assistance, and discussions of methodological issues. The authors made the final research decisions, conducted the experiments, and take responsibility for the methods, results, and claims.


\bibliography{anthology,custom}
\bibliographystyle{acl_natbib}
\newpage

\appendix

\section{Datasets Statistics}
\label{sec:appendix_datasets}
We download datasets GSM8K, GPQA, and CommonsenseQA from Huggingface, MATH from their official project website: \url{https://github.com/hendrycks/math}. GSM8K dataset is split according to the official original split ratio. We use the official training set for Math and MATH-500 for the test set due to its high representation and low cost. Since there is no official train/test split for GPQA, we use gpqa\_main and gpqa\_extended for training which excludes gpqa\_diamond to ensure a disjoint evaluation set, and gpqa\_diamond as the test set. Table~\ref{table_dataset} shows the statistics of all datasets.
\begin{table}[h]
\centering
\small
{\begin{tabular}{llcc}
\toprule Dataset & Type & \#Train & \#Test\\ \midrule
GSM8K & Mathematics & 7473  & 1319\\
MATH & Mathematics & 7500 & 500\\
GPQA & Science & 348 & 198\\
CommonsenseQA & Commonsense & 9740 & 1220 \\
 \bottomrule
\end{tabular}}
\caption{\label{table_dataset} Dataset statistics. }
\end{table}

\section{Implementation Details}
\label{sec:appendix_implement}
Primary experiments are conducted on four Nvidia A100-80GB GPUs.

\subsection{Teacher LLM Parameters}
We access DeepSeek-V3.2~\citep{deepseek2025v32} through the official \texttt{deepseek-reasoner} API endpoint: \url{https://api-docs.deepseek.com}.
\begin{table}[h]
\centering
{\begin{tabular}{lc}
\toprule Parameter & Value \\ \midrule
temperature & 0 \\
max new tokens & 2048\\
do sample & True\\
model & deepseek-reasoner \\
response\_format & \{'type': 'text'\} \\
\bottomrule
\end{tabular}}
\caption{\label{table_large_param} Teacher LLM parameter settings.}
\end{table}

\subsection{Student SLMs Parameters}
Experiments are performed with the Huggingface trl framework. We use four Nvidia A100-80GB GPUs with BF16 and LoRA~\cite{hu2022lora} for training and evaluation. The inference parameter settings across all datasets are shown in Table~\ref{table_student_infer}. The training hyperparameter settings across all datasets are shown in Table~\ref{table_sft_hyper} and Table~\ref{table_rl_hyper}. 

\begin{table}[H]
\centering
{\begin{tabular}{lc}
\toprule Parameter & Value \\ \midrule
temperature & 0.2 \\
max\_new\_tokens & 2048\\
top\_p & 0.9\\
top\_k & 50\\
do\_sample & True\\ 
batch\_size & 16 \\ \bottomrule
\end{tabular}}
\caption{\label{table_student_infer} Student SLMs inference parameter settings.}
\end{table}

\begin{table}[h]
\small
\centering
{\begin{tabular}{lc}
\toprule Hyperparameter & Value \\ \midrule
epoch & 5 \\
batch\_size & 8\\
learning\_rate & 2e-5\\
max\_seq\_length & 2048\\
gradient\_accumulation\_steps & 4\\
warmup\_ratio & 0.03\\
weight\_decay & 0.01\\
lr\_scheduler\_type & "linear" \\
max\_grad\_norm & 1.0\\ 
lora\_alpha & 16\\
lora\_dropout & 0.1\\
r & 8\\
task\_type & "CAUSAL\_LM"\\
target\_modules & ["q\_proj", "v\_proj"]\\
\bottomrule
\end{tabular}}
\caption{\label{table_sft_hyper} Student SLMs SFT hyperparameter settings.}
\end{table}

\begin{table}[h]
\small
\centering
{\begin{tabular}{lc}
\toprule Hyperparameter & Value \\ \midrule
epoch & 1 \\
batch\_size & 4\\
learning\_rate & 1e-5\\
max\_completion\_length & 2048\\
beta & 0.04\\
num\_generations & 4\\
gradient\_accumulation\_steps & 4\\
scale\_rewards & "group" \\
lora\_alpha & 32\\
lora\_dropout & 0.1\\
r & 8\\
task\_type & "CAUSAL\_LM"\\
target\_modules & ["q\_proj", "v\_proj"]\\
seed & 730, 731, 732\\
\bottomrule
\end{tabular}}
\caption{\label{table_rl_hyper} Student SLMs RL hyperparameter settings.}
\end{table}

\section{Hyperparameter Tuning}
\label{sec:appendix_tune}
We selected the process-reward weight $\lambda$ and other training hyperparameters using held-out subsets sampled from the corresponding training data. We used a 5\% held-out development split for GSM8K, MATH, GPQA, and CommonsenseQA. We tuned $\lambda$ over $\{0, 0.25, 0.5, 0.75, 1.0\}$ and selected learning rate, number of epochs, and other hyperparameters using these development subsets. The benchmark evaluation splits reported in the main tables were reserved for final evaluation and were not used for model or hyperparameter selection.

\section{Analysis of LCS-F1 Alignment}
\label{sec:appendix_lcs_norm}
SPEAR uses the LCS-F1 score to balance two complementary objectives. Alignment precision, $|\mathrm{LCS}|/|\mathcal{A}_s|$, penalizes unsupported or repetitive student anchors, whereas teacher-milestone recall, $|\mathrm{LCS}|/|\mathcal{A}_t|$, penalizes missing teacher anchors. Using either component alone leaves a degenerate strategy: precision alone can give full credit to a short matching prefix, while recall alone can give full credit to a verbose trajectory containing all teacher milestones plus unsupported anchors. Their harmonic mean penalizes both behaviors and assigns full credit only to complete, concise alignment.

The example in Table~\ref{table_norm_example} makes this balance explicit. Let the teacher trajectory be $\mathcal{A}_t=[a_1,a_2,a_3]$. Precision alone assigns the maximum score to the incomplete trajectory $[a_1]$, while recall alone assigns the maximum score to a verbose trajectory containing $[a_1,a_2,a_3]$ plus unsupported anchors. LCS-F1 reduces both scores and reserves full credit for complete alignment. Because the teacher provides one reference trajectory rather than an exhaustive set of valid solutions, however, the reward can still under-credit a correct alternative path whose extracted milestones differ from the teacher's.

\begin{table}[h]
\centering
\small
\resizebox{\columnwidth}{!}{
\begin{tabular}{lcccc}
\toprule
\textbf{Student trajectory $\mathcal{A}_s$} & $|\mathcal{A}_s|$ & Precision & Recall & LCS-F1 \\
\midrule
$[a_1,a_2,a_3]$ & 3 & 1.00 & 1.00 & 1.00 \\
$[a_1]$ & 1 & 1.00 & 0.33 & 0.50 \\
$[a_1,x,a_2,y,a_3,z]$ & 6 & 0.50 & 1.00 & 0.67 \\
$[a_1,a_2,x]$ & 3 & 0.67 & 0.67 & 0.67 \\
\bottomrule
\end{tabular}}
\caption{Illustrative behavior of precision, recall, and LCS-F1. $\mathcal{A}_t=[a_1,a_2,a_3]$. Symbols $x,y,z$ denote unsupported student anchors.}
\label{table_norm_example}
\end{table}

\section{Pseudocode for Symbolic Anchor Extraction}
\label{sec:appendix_pseudocode}
Algorithm~\ref{alg:anchor_extraction} instantiates the domain-adaptive projection $\Phi(y,\tau)$ invoked for both teacher and student responses in Algorithm~\ref{alg:spear}. It implements the extraction rules described in Section~\ref{subsec:anchors} and returns an ordered symbolic trajectory $\mathcal{A}$. The resulting teacher and student trajectories are passed to the LCS-F1 calculation in Equation~\ref{eq:lcs_f1}, which jointly penalizes missing teacher milestones and unsupported student milestones.

\begin{algorithm}[H]
\caption{Domain-Adaptive Symbolic Projection $\Phi$}
\label{alg:anchor_extraction}
\begin{algorithmic}[1]
\Procedure{$\Phi$}{$y,\tau$}
    \State $\mathcal{A} \gets [\,]$
    \If{$\tau \in \{\mathrm{GSM8K},\mathrm{MATH}\}$}
        \State $\mathcal{A} \gets$ ordered LaTeX expressions and variable assignments in $y$
        \State \Comment{Exclude bare numerical values}
    \ElsIf{$\tau = \mathrm{GPQA}$}
        \State $D \gets \Call{DependencyParse}{y}$
        \ForAll{governing verbs $v$ in $D$}
            \If{$v$ has an object $obj$}
                \State $a \gets (\mathrm{lemma}(v),\mathrm{span}(obj))$
            \ElsIf{$v$ has a subject $sbj$}
                \State $a \gets (\mathrm{span}(sbj),\mathrm{lemma}(v))$
            \Else
                \State \textbf{continue}
            \EndIf
            \If{$a \notin \mathcal{A}$}
                \State append $a$ to $\mathcal{A}$
            \EndIf
        \EndFor
    \Else
        \State $D \gets \Call{DependencyParse}{y}$
        \ForAll{noun chunks $c$ in $D$}
            \State $r \gets \mathrm{root}(c),\ h \gets \mathrm{head}(c)$
            \If{$h$ is a verb}
                \State $a \gets (r,\mathrm{lemma}(h))$
            \Else
                \State $a \gets r$
            \EndIf
            \State append $a$ to $\mathcal{A}$
        \EndFor
    \EndIf
    \State \Return $\mathcal{A}$
\EndProcedure
\end{algorithmic}
\end{algorithm}

\section{Cross-Task Applicability on AlpacaEval 2.0}
\label{sec:appendix_alpaca}
To evaluate whether our open-domain state transition extractor applies beyond commonsense reasoning, we use the same extraction rules, without task-specific modification, on the open-ended instruction-following benchmark AlpacaEval 2.0.\footnote{\url{https://github.com/tatsu-lab/alpaca_eval}} We evaluate Qwen3-4B using the official length-controlled (LC) win rate. As shown in Table~\ref{table_alpaca}, SPEAR improves the corresponding GRPO, Dr. GRPO, and DAPO baselines by 1.61\%, 1.12\%, and 1.74\%, respectively, and also outperforms the Logic-RL reward under each optimization framework. These results provide evidence that the open-domain extractor captures reusable structure beyond commonsense question answering, rather than operating as a benchmark-specific heuristic.

\begin{table}[h]
\centering
\small
\begin{tabular}{lc}
\toprule
\textbf{Qwen3-4B Method} & \textbf{LC Win Rate (\%)} \\
\midrule
Zero-shot (Base) & 49.07 \\
\midrule
GRPO & 54.41 \\
GRPO + Logic-RL & 54.78 \\
GRPO + SPEAR & \textbf{56.02} \\
\midrule
Dr. GRPO & 54.66 \\
Dr. GRPO + Logic-RL & 54.53 \\
Dr. GRPO + SPEAR & \textbf{55.78} \\
\midrule
DAPO & 55.90 \\
DAPO + Logic-RL & 56.27 \\
DAPO + SPEAR & \textbf{57.64} \\
\bottomrule
\end{tabular}
\caption{Cross-task applicability on AlpacaEval 2.0 using the unchanged open-domain state-transition extractor.}
\label{table_alpaca}
\end{table}

\section{Measured Resource and Training Overhead}
\label{sec:appendix_efficiency}
Table~\ref{table_efficiency_breakdown} reports the numerical resource comparison for Qwen3-4B on MATH using four A100-80GB GPUs.

\begin{table}[h]
\centering
\small
\begin{tabular}{lcc}
\toprule
\textbf{Metric} & \textbf{Neural PRM} & \textbf{SPEAR} \\
\midrule
Time per step (s) & 216.67 & \textbf{169.74} \\
Slowdown vs. GRPO (\%) & 30.04 & \textbf{1.87} \\
Extra neural models & 1 & \textbf{0} \\
Reward parameters (B) & 7 & \textbf{0} \\
GPU reward passes/rollout & 1 & \textbf{0} \\
Relative teacher API cost & $1\times$ & $1\times$ \\
\bottomrule
\end{tabular}
\caption{Resource overhead on Qwen3-4B MATH. Time is measured on four A100-80GB GPUs; GRPO takes 166.62 s/step.}
\label{table_efficiency_breakdown}
\end{table}

PRM scoring retains an additional 7B reward model in GPU memory and performs one extra transformer forward pass for every rollout. SPEAR introduces no neural reward model: spaCy extraction and LCS-F1 matching execute on CPU, producing negligible additional GPU memory overhead for reward scoring. Although both approaches consume the same teacher CoTs, training a new PRM generally requires separately process-labeled supervision; PRM800K, for example, contains 800K step-level annotations~\cite{lightman2024let}, whereas SPEAR derives its reward directly from the teacher trajectories.

The measured time per step increases by 30.04\% with Qwen2.5-Math-PRM but only 1.87\% with SPEAR. The complete SPEAR run took 23 hours and 42 minutes, with 76.59~GB peak memory per GPU, 83 CPU threads, and approximately 0.77\% process CPU utilization. Thus, SPEAR's principal computational advantage is eliminating neural reward-model training and per-rollout inference.

\section{Performance with Mismatched Anchor Extractors}
\label{sec:appendix_ood}
To evaluate whether SPEAR depends on matching its symbolic extractor to the task domain, we deliberately swap the extraction rules across domains: we apply the causal-dependency extractor to MATH and the mathematical-state extractor to GPQA. Table~\ref{table_ood} reports the resulting Qwen3-4B performance under these mismatched extraction settings.

Compared with the matched extractors in Table~\ref{table_results}, mismatched extraction reduces MATH accuracy by 2.47 percentage points and GPQA accuracy by 4.72 points on average across the three RL frameworks. It also falls below the corresponding outcome-only RL baseline in five of the six settings; the sole exception is DAPO on MATH, which remains 0.20 points higher. These results support our domain-adaptive design: useful symbolic supervision depends on extracting milestones appropriate to the reasoning structure of each task.

\begin{table}[htb]
\centering
\resizebox{1\columnwidth}{!}{\begin{tabular}{lll}
\toprule
Method & MATH & GPQA \\ \midrule
GRPO+SPEAR  & 71.00 & 36.36 \\
Dr. GRPO+SPEAR  & 71.60 & 36.87 \\
DAPO+SPEAR  & 73.00 & 38.38 \\
\bottomrule
\end{tabular}}
\caption{\label{table_ood}Performance of SPEAR on Qwen3-4B with deliberately mismatched anchor extractors. We apply the causal-dependency extractor to MATH and the mathematical-state extractor to GPQA.}
\end{table}

\section{Prompt Templates}
\label{sec:appendix_prompts}
\begin{itemize}

\item System prompt for all models.
\begin{itemize}
"A conversation between User and Assistant. The user asks a question, and the Assistant solves it. The assistant first thinks about the reasoning process in the mind and then provides the user with the answer. The reasoning process and answer are enclosed within <think> </think> and <answer> </answer> tags, respectively, i.e., <think> reasoning process here </think><answer> answer here </answer>."
\end{itemize}

\item Prompt templates for different benchmarks.
\begin{itemize}
\item For benchmarks GSM8K:  \\
question + "Let's think step by step." 
\item For benchmark MATH:  \\
question + "Let's think step by step and put the final answer in \verb|\boxed{}|." 
\item For benchmark GPQA:  \\
question + "Put the final answer in \verb|\boxed{}|." + options
\item For benchmark CommonsenseQA:  \\
question + " Put the final answer in \verb|\boxed{}|." + options
\end{itemize}
\end{itemize}

\section{Case Study of Distillation Impact on Student LM's Output}
\label{sec:appendix_case}
Table~\ref{table_case_math}, \ref{table_case_gpqa}, \ref{table_case_csqa} provides the comparisons of student model Llama-3-8B-Instruct behavior difference before and after the SPEAR distillation across benchmarks. It is evident that the <think> and <answer> parts from teacher LLMs help the student LM with reasoning think and finally reach the correct answer.


\begin{table*}[h]
\centering
\begin{tabular}{p{3.5cm}|p{12cm}}
\multicolumn{2}{c}{\textbf{MATH Example}} \\ \hline

Question &
Express $\frac{165_7}{11_2}+\frac{121_6}{21_3}$ in base 10. 
\textbf{Gold answer:} \textcolor{ForestGreen}{39}
\\ \hline

Student’s output before distillation &
First, convert the numbers into base 10.

$165_7 = 1 \times 7^2 + 6 \times 7 + 5 = 96$

$11_2 = 2$

So,
$\frac{165_7}{11_2} = \frac{96}{2} = 48$

Next,

$121_6 = 36 + 12 + 1 = 49$

$21_3 = 2 \times 3 + 1 = 7$

Thus,
$\frac{121_6}{21_3} = \frac{49}{7} = 7$

Adding them together gives
\textcolor{Red}{$48 + 7 = \boxed{55}$}.
\\ \hline

Teachers' response &
<think>The expression is evaluated by converting each number from its given base to base 10:

$165_7 = 1 \times 7^2 + 6 \times 7^1 + 5 \times 7^0 = 49 + 42 + 5 = 96$

$11_2 = 1 \times 2^1 + 1 \times 2^0 = 2 + 1 = 3$

$\frac{165_7}{11_2} = \frac{96}{3} = 32$

Next,
    
$121_6 = 1 \times 6^2 + 2 \times 6^1 + 1 \times 6^0 = 36 + 12 + 1 = 49$

Thus, $21_3 = 2 \times 3^1 + 1 \times 3^0 = 6 + 1 = 7$

$\frac{121_6}{21_3} = \frac{49}{7} = 7$

Adding the results: $32 + 7 = 39$.
</think>

<answer>Thus, the expression in base 10 is 
\textcolor{ForestGreen}{$\boxed{39}$}.</answer>
\\ \hline

Student’s output after distillation &
<think>
Convert each number into base 10 before evaluating the fractions.

$165_7 = 1 \times 7^2 + 6 \times 7^1 + 5 \times 7^0 = 49 + 42 + 5 = 96$

$11_2 = 1 \times 2^1 + 1 \times 2^0 = 2 + 1 = 3$

$\frac{165_7}{11_2} = \frac{96}{3} = 32$

$121_6 = 1 \times 6^2 + 2 \times 6^1 + 1 \times 6^0 = 36 + 12 + 1 = 49$

$21_3 = 2 \times 3^1 + 1 \times 3^0 = 6 + 1 = 7$

$\frac{121_6}{21_3} = \frac{49}{7} = 7$

Finally, $32 + 7 = 39$
</think>

<answer>
Therefore, the expression in base 10 is
\textcolor{ForestGreen}{$\boxed{39}$}.</answer>
\\ \hline
\end{tabular}
\caption{\label{table_case_math} Case study of student LM Llama-3-8B-Instruct output on MATH before and after distillation.}
\end{table*}

\begin{table*}[h]
\centering
\begin{tabular}{p{3.5cm}|p{12cm}}
\multicolumn{2}{c}{\textbf{GPQA Example}} \\ \hline

Question &
What is the major outcome of the reaction between cyclopent-1-enol and bromine?

\textbf{Options:} \textcolor{ForestGreen}{2-bromocyclopentanone}; \textcolor{Red}{3-bromocyclopentanone};

\textcolor{Red}{(1R,2R)-1,2-dibromocyclopentanol};
\textcolor{Red}{(1R,2S)-1,2-dibromocyclopentanol}
\\ \hline

Student’s output before distillation &
Bromine usually adds across double bonds in alkenes. Since cyclopent-1-enol contains a double bond, bromine reacts by forming a dibrominated product across the alkene.

Therefore, the major product is 
\textcolor{Red}{(1R,2R)-1,2-dibromocyclopentanol}.
\\ \hline

Teachers' response &
<think>
Cyclopent-1-enol is an enol, meaning it contains both a carbon-carbon double bond and an alcohol group directly attached to the alkene carbon. Enols are generally unstable and readily undergo keto-enol tautomerization.

Under bromination conditions, the enol form reacts as a nucleophilic alkene. Bromine electrophilically attacks the double bond, generating an $\alpha$-brominated carbonyl compound after tautomerization.

First, cyclopent-1-enol tautomerizes to cyclopentanone:

\[
\text{cyclopent-1-enol} \rightleftharpoons \text{cyclopentanone}
\]

Then, bromination occurs at the $\alpha$-position adjacent to the carbonyl group through enolization of cyclopentanone.

This produces the $\alpha$-brominated ketone:

\[
\text{cyclopentanone} + Br_2 \rightarrow \text{2-bromocyclopentanone}
\]

Therefore, the reaction does not mainly give dibromination across the double bond. Instead, the major product is the brominated ketone formed via keto-enol tautomerization and $\alpha$-substitution chemistry.
</think>

<answer>
Thus, the major product of the reaction is 
\textcolor{ForestGreen}{2-bromocyclopentanone}.
</answer>
\\ \hline

Student’s output after distillation &
<think>
Cyclopent-1-enol is an enol, which can tautomerize into cyclopentanone. In the presence of bromine, ketones typically undergo $\alpha$-bromination at the carbon adjacent to the carbonyl group.

After tautomerization to cyclopentanone, bromination occurs at the $\alpha$-position, producing the brominated ketone.
</think>

<answer>
Therefore, the major product is 
\textcolor{ForestGreen}{2-bromocyclopentanone}.
</answer>
\\ \hline
\end{tabular}
\caption{\label{table_case_gpqa} Case study of student LM Llama-3-8B-Instruct output on GPQA before and after distillation.}
\end{table*}


\begin{table*}[t]
\centering
\begin{tabular}{p{3.5cm}|p{12cm}}
\multicolumn{2}{c}{\textbf{CommonsenseQA Example}} \\ \hline

Question &
What is a great place to lay in the sun? 

\textbf{Options:} 
\textcolor{Red}{A. in the basement};
\textcolor{Red}{B. west};
\textcolor{Red}{C. solar system};
\textcolor{ForestGreen}{D. beach};
\textcolor{Red}{E. beans}
\\ \hline

Student’s output before distillation &
A good place to lay in the sun should be somewhere related to sunlight or space. The solar system contains the sun and receives sunlight directly.

Therefore, the best answer is 
\textcolor{Red}{C. solar system}.
\\ \hline

Teachers' response &
<think>
The question asks for a place where people commonly relax or sunbathe.

Evaluate each option:

- A. in the basement: basements are underground and usually dark, so they are not good places to lay in the sun.

- B. west: west is a direction, not a physical place where someone would lay down.

- C. solar system: although the sun is part of the solar system, people cannot physically lay there. This option is too broad and unrealistic.

- D. beach: beaches are commonly associated with relaxing, sunbathing, and laying in the sun. People frequently go to beaches specifically to enjoy sunlight.

- E. beans: beans are food and not a place.

Among all options, "beach" is the only realistic and commonsense location associated with laying in the sun.
</think>

<answer>
Therefore, the correct answer is 
\textcolor{ForestGreen}{D. beach}.
</answer>
\\ \hline

Student’s output after distillation &
<think>
The question asks for a realistic place where people commonly lay in the sun.

A basement is dark, west is only a direction, the solar system is not a practical location, and beans are food. A beach is a common place for relaxing and sunbathing.
</think>

<answer>
Thus, the correct answer is 
\textcolor{ForestGreen}{D. beach}.
</answer>
\\ \hline
\end{tabular}
\caption{\label{table_case_csqa} Case study of student LM Llama-3-8B-Instruct output on CommonsenseQA before and after distillation.}
\end{table*}

\end{document}